\documentclass[runningheads]{llncs}
\usepackage[T1]{fontenc}
\usepackage{graphicx}
\usepackage{color}
\usepackage{algorithm,algorithmic}
\usepackage{mathtools}
\usepackage{mathptmx}
\usepackage{amsmath}
\usepackage{amssymb}
\usepackage{url}

\begin{document}
\title{TinyTNAS: GPU-Free, Time-Bound, Hardware-Aware Neural Architecture Search for TinyML Time Series Classification }
\titlerunning{TinyTNAS}
%
\author{Bidyut Saha \and
Riya Samanta \and
Soumya K. Ghosh \and  Ram Babu Roy}
\authorrunning{B. Saha et al.}
%
\institute{Indian Institute of Technology Kharagpur, India \\
\email{bidyutsaha@kgpian.iitkgp.ac.in, riya.samanta@iitkgp.ac.in, skg@cse.iitkgp.ac.in, rambabu@see.iitkgp.ac.in}}
\maketitle              
\begin{abstract}

In this work, we present TinyTNAS, a novel hardware-aware multi-objective Neural Architecture Search (NAS) tool specifically designed for TinyML time series classification. Unlike traditional NAS methods that rely on GPU capabilities, TinyTNAS operates efficiently on CPUs, making it accessible for a broader range of applications. Users can define constraints on RAM, FLASH, and MAC operations to discover optimal neural network architectures within these parameters. Additionally, the tool allows for time-bound searches, ensuring the best possible model is found within a user-specified duration. By experimenting with benchmark datasets—UCI HAR, PAMAP2, WISDM, MIT-BIH, and PTB Diagnostic ECG Database—TinyTNAS demonstrates state-of-the-art accuracy with significant reductions in RAM, FLASH, MAC usage, and latency. For example, on the UCI HAR dataset, TinyTNAS achieves a 12x reduction in RAM usage, 144x reduction in MAC operations, and 78x reduction in FLASH memory while maintaining superior accuracy and reducing latency by 149x. Similarly, on the PAMAP2 and WISDM datasets, it achieves a 6x reduction in RAM usage, 40x reduction in MAC operations, an 83x reduction in FLASH, and a 67x reduction in latency, all while maintaining superior accuracy. Notably, the search process completes within 10 minutes in a CPU environment. These results highlight TinyTNAS's capability to optimize neural network architectures effectively for resource-constrained TinyML applications, ensuring both efficiency and high performance.  The code for TinyTNAS is available at the GitHub repository and can be accessed through \url{https://github.com/BidyutSaha/TinyTNAS.git}.

\end{abstract}
\section{Introduction}

Neural Architecture Search (NAS) offers significant advantages, including automating the design of neural network architectures, thereby saving considerable time and expert effort required in manual design \cite{chauhan2023dqnas}. It also enhances performance by exploring and optimizing a wide range of architectures that might not have been feasible through manual methods, leading to more efficient and effective neural networks \cite{xie2023architecture}. Traditionally, NAS relies on extensive computational resources, predominantly GPUs, to explore vast search spaces and identify optimal configurations. This involves determining the best combination of layers, operations, and hyperparameters to maximize model performance for specific tasks \cite{zoph2016neural}. While effective, these methods are resource-intensive and often impractical for applications requiring rapid deployment or operating within hardware constraints. In response to the limitations of traditional NAS, \textit{hardware-aware NAS} methods have been developed. These approaches integrate hardware parameters such as memory usage, computational complexity (measured in FLOPs or MACs), and power consumption into the architecture search process \cite{cai2018proxylessnas}. The hardware-aware NAS ensures that the resulting models are not only high-performing but also suitable for deployment on resource-constraint devices, such as microcontrollers (MCUs) and edge devices. This is particularly important for applications in the Internet of Things (IoT), where computational resources are limited. \textit{Multi-objective NAS} further extends this concept by simultaneously optimizing for multiple criteria, such as accuracy, latency, and resource usage \cite{dong2018dpp}. This approach balances trade-offs between different objectives, enabling the discovery of architectures that meet diverse and sometimes conflicting requirements. Multi-objective optimization is crucial for developing models that are both efficient and effective across various deployment scenarios.

\begin{figure}[!t]
 \centering
  \includegraphics[width=\linewidth]{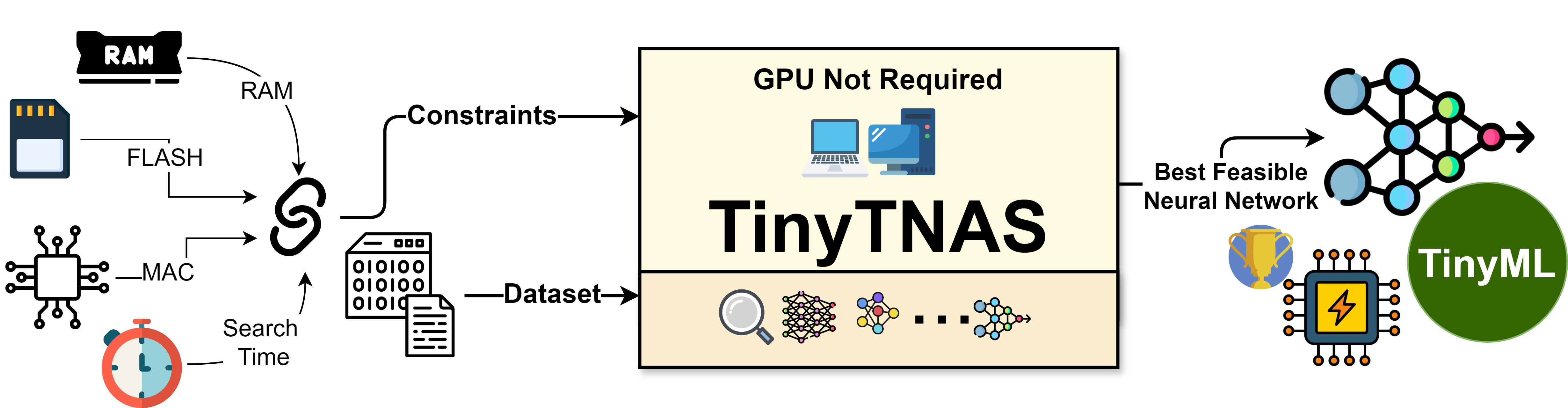} 
  \vspace{-0.1in}
  \caption{Concept Diagram of TinyTNAS}
  \label{concept}
  \vspace{-0.1in}
\end{figure}

As the demand for intelligent edge devices grows, there is an increasing need to deploy machine learning models on MCU and other resource-constraint hardware. This is where TinyML comes into play. TinyML represents the intersection of machine learning and embedded systems, focusing on deploying machine learning models on MCUs and other resource-constraint devices \cite{warden2019tinyml,saha2023bandx}. The primary motivation for TinyML is to bring intelligence to edge devices, enabling real-time decision-making and reducing reliance on cloud-based processing. This shift offers several advantages, including lower latency, reduced bandwidth usage, and enhanced privacy since data can be processed locally \cite{lane2016deepx}. However, developing TinyML solutions presents significant challenges. The limited computational power, memory, and storage available on MCUs necessitate highly efficient models \cite{david2021tensorflow}. Traditional approaches to model optimization for TinyML often involve manual tuning and simplification of pre-designed architectures, which is both time-consuming and sub-optimal. Moreover, balancing performance with resource constraints requires a deep understanding of both machine learning and embedded system design. Despite these challenges, there have been notable successes in the TinyML domain. For instance, models optimized for voice recognition and keyword spotting have been successfully deployed on MCUs, enabling applications like always-on voice assistants \cite{banbury2021micronets}. Similarly, accelerometer-based activity recognition systems have been implemented on wearable devices, demonstrating the potential of TinyML for fitness tracking, health monitoring and human computer interactions \cite{saha2023bandx,saha2023tinyml,saha2023wrist,saha2024personalized}.

Integrating NAS with TinyML offers a promising solution to overcome the challenges of model optimization for resource-constraint environments. By leveraging NAS, it is possible to automate the design of efficient neural network architectures tailored to the specific hardware constraints of TinyML devices. This approach can significantly accelerate the development process and ensure optimal performance \cite{he2021automl}. Despite the advancements in NAS and TinyML, there are still significant challenges in achieving state-of-the-art (SOTA) performance while adhering to strict resource constraints. Traditional NAS methods are typically impractical  due to their reliance on GPUs and extensive computational requirements. Therefore, there is a critical need for NAS methods that can operate efficiently on CPUs and deliver high-quality models within a reasonable time frame.

In this context, we introduce \textbf{TinyTNAS}\footnote{The code repository is available at \url{https://github.com/BidyutSaha/TinyTNAS.git}.}, a novel \textit{hardware-aware multi-objective NAS tool} specifically designed for \textit{TinyML time series classification} (see Figure \ref{concept}). Unlike existing methods, TinyTNAS operates efficiently on CPUs, making it accessible and practical for a broader range of applications. Users can define constraints on RAM, FLASH, and MAC operations to discover optimal neural network architectures that meet these parameters. Additionally, TinyTNAS allows for \textit{time-bound} searches, ensuring the best possible model is found within a user-specified duration. By experimenting with benchmark datasets—UCIHAR \cite{misc_human_activity_recognition_using_smartphones_240}, PAMAP2 \cite{misc_pamap2_physical_activity_monitoring_231}, WISDM \cite{misc_wisdm_smartphone_and_smartwatch_activity_and_biometrics_dataset__507}, MIT-BIH \cite{moody2001impact}, and {PTB Diagnostic ECG Database \cite{bousseljot1995nutzung}—TinyTNAS demonstrates state-of-the-art accuracy with significant reductions in RAM, FLASH, and MAC usage.

Our objective is to provide a comprehensive solution that bridges the gap between NAS and TinyML, enabling the efficient deployment of high-accuracy models on resource-constraint devices. TinyTNAS represents a substantial advancement in the TinyML domain, offering a practical tool for optimizing neural networks within stringent hardware constraints and delivering superior performance in time-sensitive environments.

\section{Related Work}

Neural Architecture Search (NAS) has revolutionized the automated design of neural networks, significantly enhancing performance and reducing the need for manual intervention. Traditional NAS methods, such as \cite{zoph2016neural} and  \cite{real2019regularized}, have primarily relied on extensive GPU resources to explore vast search spaces, resulting in high-performing but computationally expensive models. These methods often overlook the constraints of deploying models on resource-limited devices like microcontrollers (MCUs) used in TinyML applications. Recent advancements in hardware-aware NAS (HW NAS), exemplified by works like \cite{wu2019fbnet}and \cite{cai2019once}, integrate hardware metrics such as memory usage and computational complexity into the search process, producing models that are more suitable for deployment on edge devices. However, these approaches still predominantly rely on GPUs for model optimization, limiting their accessibility and practicality for broader applications.

In the domain of TinyML, recent works like MCUNet \cite{lin2020mcunet} and MicroNets \cite{banbury2021micronets} have made significant strides. MCUNet introduces a joint design framework that optimizes both the neural architecture and the inference engine to fit the resource constraints of MCUs, achieving impressive efficiency and accuracy. However, MCUNet and similar tools typically require heavy GPU resources for their computation, making them less feasible for deployment on CPU-only environments. Similarly, MicroNets focuses on building extremely compact and efficient neural networks suitable for MCUs by leveraging advanced pruning and quantization techniques. Notably, MicroNets also faces challenges with heavy GPU requirements, limiting its practicality in CPU-centric environments. Despite these advancements, a significant research gap remains in the specific optimization of NAS for time series classification on resource-constraint devices. Existing works often focus on image classification tasks, leaving time series data underexplored in CPU-centric environments.

Moreover, recent research \cite{garavagno2023hardware} presented an HW NAS approach that can run on CPUs, producing tiny convolutional neural networks (CNNs) targeting low-end microcontrollers. However, their approach has notable limitations, such as using standard CNN layers when more computationally efficient layers could be employed. Additionally, their method relies on traditional grid search, which is computationally expensive, and focuses on image classification rather than time series data, without incorporating time-bound searches. This restricts its applicability for tasks requiring efficient temporal data processing and adherence to specified search durations.

Our work addresses these gaps by introducing TinyTNAS, a CPU-efficient, hardware-aware NAS tool specifically designed for TinyML time series classification. TinyTNAS optimizes neural network architectures within RAM, FLASH, and MAC constraints and ensures rapid deployment by completing searches within a user defined time window on CPUs, setting a new standard for efficiency in the TinyML domain. We compare the models generated by TinyTNAS against state-of-the-art (SOTA) models for benchmark datasets—UCI HAR \cite{misc_human_activity_recognition_using_smartphones_240}, PAMAP2 \cite{misc_pamap2_physical_activity_monitoring_231}, WISDM \cite{misc_wisdm_smartphone_and_smartwatch_activity_and_biometrics_dataset__507}, MIT-BIH \cite{moody2001impact}, and PTB Diagnostic ECG Database \cite{bousseljot1995nutzung}—using existing methods known for achieving SOTA accuracy: CNN (1D) \cite{yang2015deep}, DeepConvLSTM (1D) \cite{ordonez2016deep}, and LSTM-CNN (1D) \cite{xia2020lstm} for HAR, and CNN (1D) \cite{kachuee2018ecg} for MIT-BIH and PTB-DED.

\section{TinyTNAS}
\subsection{Search Space}

In Neural Architecture Search (NAS), the \textit{search space} refers to the complete set of possible neural network architectures explored by the algorithm. NAS encompasses three primary search spaces: layer-wise, cell-wise, and hierarchical. \textit{Layer-wise NAS} focuses on optimizing individual layer configurations such as convolutional and pooling layers \cite{zoph2016neural}. \textit{Cell-wise NAS} identifies optimal repeating cell structures, enhancing scalability and efficiency \cite{zoph2018learning,liu2018darts}. \textit{Hierarchical NAS} integrates layer-wise and cell-wise strategies across multiple abstraction levels for adaptive model design \cite{liu2018progressive,cai2018proxylessnas}.
\\
\textbf{TinyTNAS} utilizes a cell-wise search space. Each generated neural network architecture follows the template, depicted in Figure \ref{searchspace}.

\begin{itemize}
  \item Candidate architecture begins with a Depthwise Separable 1d Convolutional layer with stride=1, kernel size=3, and \( k \) number of kernels with ReLU activation.
  \item Followed by a block that may repeat 0 to \( c \) times, where \( c \) is the maximum number of possible MaxPooling1D layers of size=2 adaptable to the input shape from the dataset.
  \item Next a GlobalAveragePooling1D layer.
  \item Then a Dense layer with ReLU activation and \( k_{c+1} \) neurons.
  \item Finally, a classifier layer with Dense neurons corresponding to the number of classes and a softmax function.
\end{itemize}

The repeating block structure starts with MaxPooling1D size=2  followed by Depthwise Separable 1D Convolution with kernel size=3, stride=1, ReLU activation, and filter size \( k_{i} \), where \( k_{i} \) follows a growth pattern. Specifically, \( k_{i} = 1.5 \times k_{i-1} \), starting with \( k_{1} = 1.5 \times k \). The search space is constraint by user-defined hardware limits for RAM, FLASH, and MAC. The generated model is optimised by TensorFlow Lite using \textit{integer quantization}, and its resourse requirement is estimated by MLTK profiler.

\begin{figure}[!t]
 \centering
  \includegraphics[width=\linewidth]{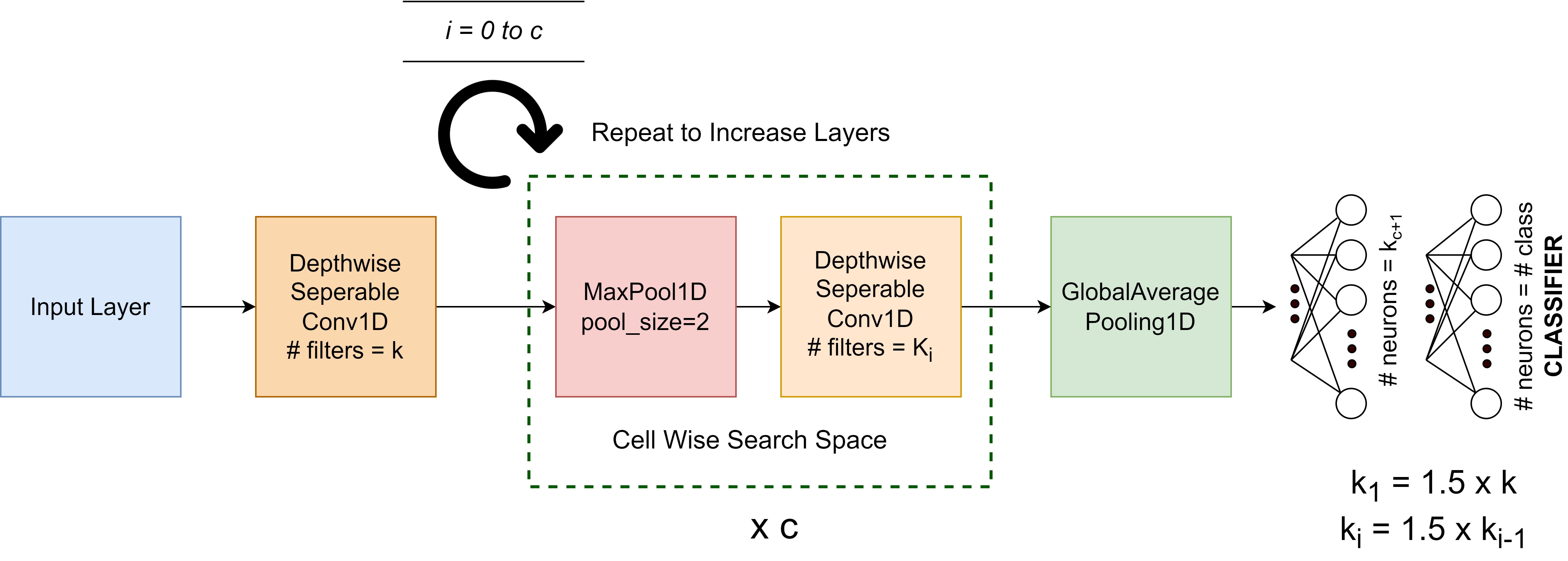} 
  \vspace{-0.1in}
  \caption{Graphical representation of candidate architectures created by TinyTNAS, where \(k\) denotes the number of filters in the first depthwise separable 1D convolutional layer and \(c\) represents the number of repeating blocks. The kernel size of each depthwise separable 1D CNN is 3 with a stride of 1 and ReLU activation. The final dense layer uses a softmax activation function, while the preceding dense layer uses a ReLU activation function.}
  \label{searchspace}
  \vspace{-0.1in}
\end{figure}

\subsection{Search Algorithm}
Our focus is on developing the TinyTNAS tool, which operates efficiently on CPUs, eliminating the need for a GPU, and generates architectures within a feasible time frame. Therefore, we avoid reinforcement learning or evolutionary algorithms and instead use a variant of grid search methods. Users can define constraints on RAM, FLASH, and MAC operations to discover optimal neural network architectures that meet these parameters. Additionally, TinyTNAS allows for time-bound searches, ensuring the best possible model is found within a user-specified duration.

Our proposed search algorithm, described in the Algorithm section, optimizes two dimensions to find the optimal architecture: \( k \) and \( c \). Here, \( k \) represents the number of filters in the first depthwise 1d convolutional layer, and \( c \) is the number of repeating blocks. The structure of the neural architecture based on \( k \) and \( c \) is detailed in the previous section and illustrated in Figure \ref{architectures}.

Unlike traditional grid search, which explores every point in the grid for \( k \) and \( c \), our method reduces computational demand by generally doubling \( k \) and keeping \( c \) constant unless the accuracy drops in consecutive explored architectures. For a detailed workflow of our search algorithm, readers are referred to Algorithm \ref{algo1}.

It is important to note that, unlike many traditional methods, our candidate architectures are trained directly on the full target dataset for four epochs. Users can adjust this value: increasing it will require more search time but potentially yield better decisions, while decreasing it will reduce search time but may lead to suboptimal decisions. We determined four epochs through empirical analysis to balance this trade-off effectively.

\begin{algorithm} [!ht]
 \caption{TinyTNAS Main Module} \label{algo1}
 \algsetup{linenosize=\tiny}
  \scriptsize
 \begin{algorithmic}[1]
 \renewcommand{\algorithmicrequire}{\textbf{Input:}}
 \renewcommand{\algorithmicensure}{\textbf{Output:}}
 \REQUIRE $RAM_{max}$, $MAC_{max}$, $Flash_{max}$, $search\_time$, dataset $ds$
 \ENSURE  First layer 1d CNN filter size $K$, Repeated block count $C$,
 \\
\noindent\textit{Initialization}: Initialize $max\_acc\_found \gets 0$, $pendings \gets [\,]$, $epoch \gets 4$, $K \gets 4$, $k \gets 4$, $C \gets 3$, $c \gets 3$
\STATE Start
\WHILE{True}
\IF{\textit{has\_search\_time}}
\STATE $model, ram, flash, mac \gets \textit{buildModel}(k, c, ds)$
\STATE $\textit{acc} \gets 0$
\IF{\textit{checkFeasibility}(\textit{$ram$, $mac$, $flash$, $RAM_{max}$, $MAC_{max}$,  $Flash_{max}$})}
\STATE $\textit{acc} \gets \textit{trainModel}(\textit{model, ds, epoch})$ \COMMENT{TensorFlow training}
\IF{\textit{max\_acc\_found} < \textit{acc}}
\STATE $\textit{max\_acc\_found} \gets \textit{acc}$
\STATE $K, k, C, pendings \gets$ \textit{updateStatus}($k, c, pendings$)
\STATE \textbf{continue}
\ENDIF
\ENDIF
\STATE \textit{is\_continueable}, $k,c$, $K$, $C \gets$\textit{exploreDepth}(\textit{$acc$, $k$, $c$, $RAM_{max}$, $MAC_{max}$,  $Flash_{max}$, $pendings$,  $search\_time$,$K$, $C$})
\IF{is\_continueable}
\STATE \textbf{continue}
\ELSE
\RETURN $K, C$
\ENDIF
\ELSE 
\RETURN $K, C$
\ENDIF
\ENDWHILE
\end{algorithmic}
\end{algorithm}

\begin{algorithm} [!ht]
 \caption{updateStatus} \label{algo2}
 \algsetup{linenosize=\tiny}
  \scriptsize
 \begin{algorithmic}[1]
 \renewcommand{\algorithmicrequire}{\textbf{Input:}}
 \renewcommand{\algorithmicensure}{\textbf{Output:}}
 \REQUIRE First layer 1d CNN filter size $k$, Repeated block count $c$, $pendings$
 \ENSURE  $K$, $k$, $C$, $pendings$
 \\
\noindent\textit{Initialization}: Initialize $\textit{multiplier} \gets 2$, $\textit{divider} \gets 4$
\STATE Start
\STATE $\textit{delta} \gets k \times \textit{multiplier} - k$
\STATE $\textit{incr} \gets \lfloor \frac{\textit{delta}}{\textit{divider}} \rfloor$
\STATE $\textit{pendings} \gets [\,]$
\IF{$\textit{incr} \geq 1$}
        \FOR{$i \gets 1$ to $\textit{divider}$}
            \STATE Push $(k + i \times \textit{incr}, c)$ to $\textit{pendings}$
        \ENDFOR
    \ENDIF
    \STATE $K \gets k$, $C \gets c$
    \STATE $k \gets k \times \textit{multiplier}$
    \RETURN $K$, $k$, $C$, $pendings$
\end{algorithmic}
\end{algorithm}

\begin{algorithm} [!ht]
 \caption{checkFeasibility} \label{algo3}
 \algsetup{linenosize=\tiny}
  \scriptsize
 \begin{algorithmic}[1]
 \renewcommand{\algorithmicrequire}{\textbf{Input:}}
 \renewcommand{\algorithmicensure}{\textbf{Output:}}
 \REQUIRE $ram$, $mac$, $flash$,  $RAM_{max}$, $MAC_{max}$,  $Flash_{max}$
 \ENSURE  $bool$
 \\
\STATE Start
\RETURN $(\textit{ram} \leq RAM_{max}) \land (\textit{flash} \leq Flash_{max}) \land (\textit{mac} \leq MAC_{max})$
\end{algorithmic}
\end{algorithm}

\begin{algorithm} [!ht]
 \caption{buildModel} \label{algo4}
 \algsetup{linenosize=\tiny}
  \scriptsize
 \begin{algorithmic}[1]
 \renewcommand{\algorithmicrequire}{\textbf{Input:}}
 \renewcommand{\algorithmicensure}{\textbf{Output:}}
 \REQUIRE First layer 1d CNN filter size $k$, Repeated block count $c$, Dataset $ds$
 \ENSURE  $model, ram, flash, mac$
 \\
\STATE Start
\STATE $model \gets$ Generate the model based on $k,c,ds$ using the method shown in Figure \ref{searchspace}.
\STATE $model^{'} \gets$ Optimise the $model$ using Integer Quantization \COMMENT{using TensorFlow Lite }
\STATE $ram, flash, mac \gets$ Profile ($model^{'})$ \COMMENT{Using MLTK library}
\RETURN $model, ram, flash, mac$
\end{algorithmic}
\end{algorithm}

\begin{algorithm} [!ht]
 \caption{exploreDepth} \label{algo5}
 \algsetup{linenosize=\tiny}
  \scriptsize
 \begin{algorithmic}[1]
 \renewcommand{\algorithmicrequire}{\textbf{Input:}}
 \renewcommand{\algorithmicensure}{\textbf{Output:}}
 \REQUIRE Accuracy $acc$, First layer 1d CNN filter size $k$, Repeated block count $c$, $RAM_{max}$, $MAC_{max}$,  $Flash_{max}$, $pendings$, $search\_time$, $K$, $C$
 \ENSURE \textit{is\_continueable}, $k,c$,
 \\
\noindent\textit{Initialization}: Initialize $\textit{accs} \gets [0]$, $\textit{cs} \gets [0]$ , $\textit{is\_continueable} \gets \textbf{False}$, $n \gets \textit{maximum\_possible\_repeatable\_blocks}$
\STATE Start
\FOR{$i \gets 0$ to $n$}
\IF{$has\_search\_time$}
\STATE $model, ram, flash, mac \gets \textit{buildModel}(k, i, ds)$
\IF{\textit{checkFeasibility}(\textit{$ram$, $mac$, $flash$, $RAM_{max}$, $MAC_{max}$,  $Flash_{max}$})}
            \STATE $\textit{acc} \gets \textit{trainModel}(\textit{model, ds, epoch})$
            \STATE Append $\textit{acc}$ to $\textit{accs}$
            \STATE Append $i$ to $\textit{cs}$
\ELSE 
            \STATE \textbf{break}
        \ENDIF
        
\ELSE
\STATE \textbf{break}
\ENDIF
\ENDFOR
\STATE $\textit{indx} \gets \textit{index of maximum value in accs}$
\IF{\textit{accs[indx]} $>$ \textit{acc}}
\STATE $c \gets \textit{cs[indx]}$
\STATE $K, k, C, pendings \gets$ \textit{updateStatus}($k, c, pendings$)
\STATE $\textit{is\_continueable} \gets \textbf{True}$
\ELSIF{\textit{pendings}}
\STATE $k, c \gets \textit{pop from pendings}$
\STATE $\textit{is\_continueable} \gets \textbf{True}$
\ENDIF
\RETURN \textit{is\_continueable}, $k,c$, $K$, $C$
\end{algorithmic}
\end{algorithm}

\section{Experiments}
\subsection{Datasets}
In our experiments, we utilize five benchmark datasets spanning the domains of lifestyle, healthcare, and human-computer interaction. Human activity recognition directly impacts lifestyle and human-computer interaction domains, and indirectly contributes to healthcare. For this purpose, we employ UCIHAR, PAMAP2, and WISDM datasets. Additionally, in healthcare applications, we leverage ECG benchmark datasets such as MIT-BIH and PTB Diagnostic ECG Database. Details for each dataset are provided below. TinyTNAS is designed to generalize across various types of time-series datasets beyond those specifically mentioned.

\begin{itemize}
    \item \textbf{\small UCIHAR :} The UCI Human Activity Recognition \cite{misc_human_activity_recognition_using_smartphones_240} dataset  captures daily activities using a waist-mounted smartphone with inertial sensors. It includes recordings of 6 activities (WALKING, WALKING\_UPSTAIRS, WALKING\_DOWNSTAIRS, SITTING, STANDING, LAYING) from 30 subjects aged 19-48. The dataset includes 3-axial linear acceleration and 3-axial angular velocity data captured at 50Hz and processed into fixed-width sliding windows of 2.56 seconds with 50\% overlap. Features were extracted from both time and frequency domains after noise filtering and gravitational separation using a low-pass filter with a cutoff frequency of 0.3 Hz.

    \vspace{0.1in}
    \item \textbf{PAMAP2 :} The PAMAP2 \cite{misc_pamap2_physical_activity_monitoring_231} dataset captures data from 18 physical activities performed by 9 subjects using 3 IMU sensors and a heart rate monitor. For simplicity, we consider 6 activities: WALKING, RUNNING, CYCLING, COMPUTER\_WORK, CAR\_DRIVING, and  ROPE\_JUMPING. Data from the IMU attached to the wrist is utilized, with sensors sampling at 100Hz originally. We resample this data to 20Hz and apply a sliding window of 2 seconds with 50\% overlap for data processing, focusing on accelerometer and gyroscope readings. This dataset is suitable for developing algorithms in data processing, segmentation, feature extraction, and activity classification.

    \vspace{0.1in}
    \item \textbf{WISDM : } The WISDM \cite{misc_wisdm_smartphone_and_smartwatch_activity_and_biometrics_dataset__507} dataset contains accelerometer and gyroscope time-series sensor data collected from a smartphone and smartwatch as 51 test subjects perform 18 activities for 3 minutes each. The raw sensor data, sampled at 20Hz, is gathered from both devices. Specifically, data from the smartwatch's accelerometer and gyroscope are used. Instead of using the recommended 10-second time window, we opt for a 2-second window with 50\% overlap to generate sliding window features, aiming to reduce computational overhead. For simplicity, the analysis focuses on six activities: WALKING, JOGGING, TYPING, WRITING, STAIRS, and BRUSHING\_TEETH.

    \vspace{0.1in}
    \item \textbf{MIT-BIH :} The MIT-BIH Arrhythmia Database \cite{moody2001impact} contains 48 half-hour excerpts of two-channel ambulatory ECG recordings from 47 subjects studied between 1975 and 1979. These recordings were digitized at 360 samples per second per channel with 11-bit resolution over a 10 mV range. The database includes approximately 110,000 beats annotated by multiple cardiologists, covering both common and less frequent clinically significant arrhythmias. In our study, we used ECG lead II data resampled to a sampling frequency of 125Hz as input, following the approach described by \cite{kachuee2018ecg}. Each beat in the dataset is annotated by at least two cardiologists and classified into five categories (N, S, V, F, Q) following the Association for the Advancement of Medical Instrumentation (AAMI) EC57 standard \cite{AAMI1998}. For preprocessing the ECG beats, we employed the pipeline described by \cite{kachuee2018ecg}.

    \vspace{0.1in}
    \item \textbf{PTB Diagnostic ECG Database :} The PTB Diagnostics dataset \cite{bousseljot1995nutzung} (PTB-DED) comprises ECG records from 290 subjects, including 148 diagnosed with MI, 52 healthy controls, and others diagnosed with 7 different diseases. Each record includes ECG signals from 12 leads sampled at 1000Hz. In our study, we focused solely on ECG lead II and analyzed the MI and healthy control categories. For preprocessing the ECG beats, we followed the pipeline detailed by \cite{kachuee2018ecg}.
\end{itemize}

\subsection{Specifications of Generated Architectures}
This paper introduces TinyTNAS, a novel hardware-aware multiobjective Neural Architecture Search (NAS) tool tailored for generating architectures under stringent resource constraints.  For our experiments, we configured TinyTNAS to operate within the limitations of \textbf{20 KB RAM}, \textbf{64 KB FLASH} memory, \textbf{60K Multiply-Accumulate Operations (MAC)}, and a maximum search time of \textbf{10 minutes} per dataset.

The tool was deployed on a desktop system equipped with an \textbf{AMD Ryzen 5 Pro 4650G processor, 32 GB of RAM, and a 256 GB SSD}, leveraging \textbf{CPU computation} with the absence of a GPU. We uniformly applied these constraints across the aforementioned five distinct datasets, utilizing TinyTNAS to discover optimal architectures.  Figure \ref{architectures} in our study visualizes the architectures identified along with corresponding search times.

\begin{figure}[ht]
 \centering
  \includegraphics[width=\linewidth]{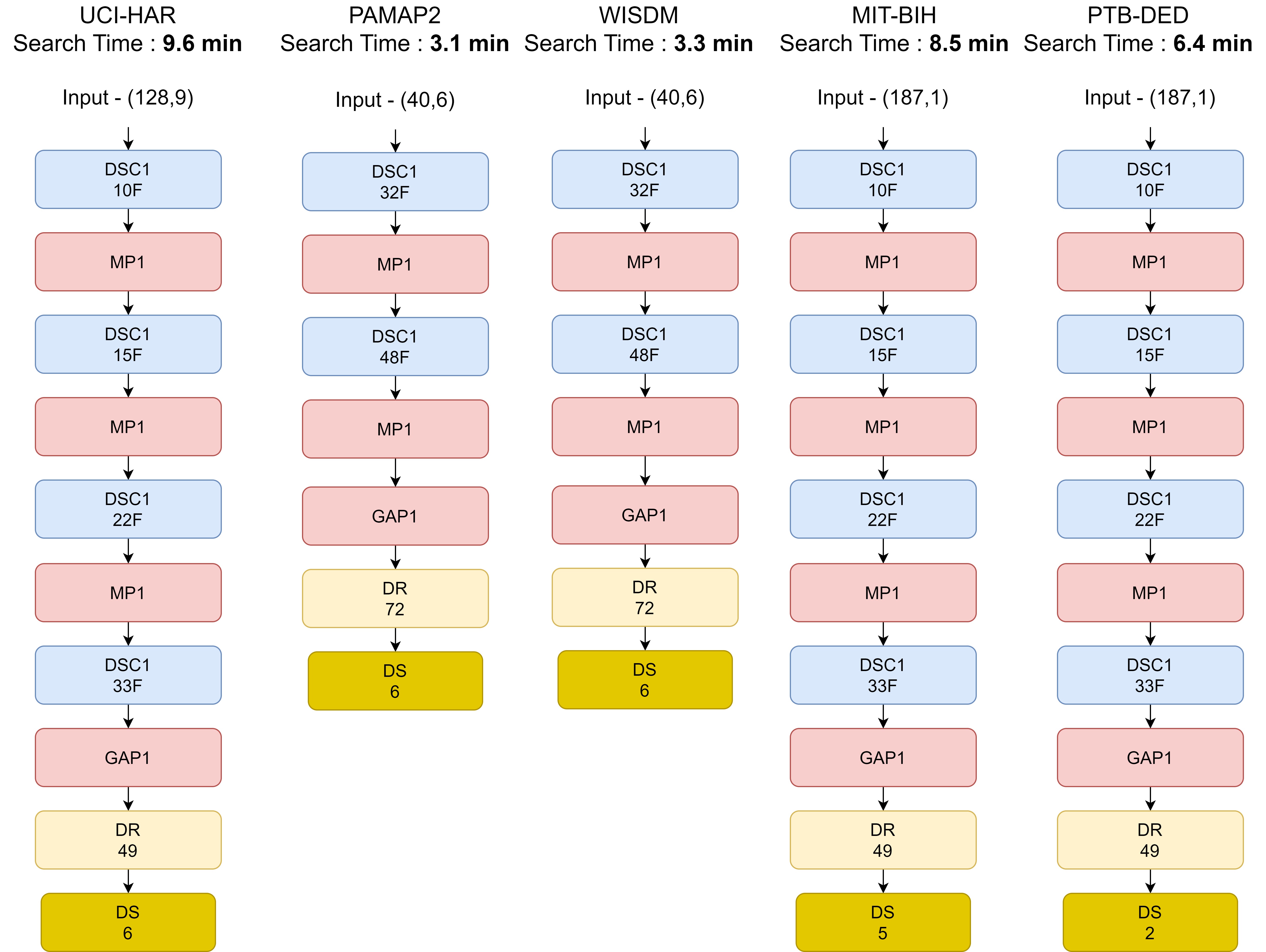} 
  \vspace{-0.1in}
  \caption{Architectures Generated by TinyTNAS under Specific Constraints on Various Datasets. Constraints include maximum RAM of 20 KB, maximum FLASH of 64 KB, maximum MAC of 60K, and a maximum search time of 10 minutes. DSC1 denotes Depthwise Separable 1D Convolution with a kernel size of 3 and ReLU activation. MP1 represents Max Pooling 1D with a size of 2. GAP1 indicates Global Average Pooling 1D. DR refers to a Dense Layer with ReLU activation, and DS denotes a Dense Layer with Softmax activation.}
  \label{architectures}
\end{figure}

\subsection{Comparision with State-Of-The-Art (SOTA)}
\vspace{-0.05in}
To evaluate the architecture generated by TinyTNAS under specific constraints with the dataset, we conduct full training using the \textit{Adam optimizer} with a learning rate of $0.001$. During training, we utilize the \textit{ReduceLROnPlateau} callback with monitoring of maximum validation accuracy, ensuring improved model convergence and efficiency. The best model, based on \textit{maximum validation accuracy}, is saved periodically in a .h5 file for future use every $200$ epochs.

For the Human Activity Recognition (HAR) dataset, we benchmark the performance of our optimized architecture against SOTA methods including \textit{CNN-based} \cite{yang2015deep}, \textit{Deep-Conv-LSTM-based} \cite{ordonez2016deep}, and \textit{LSTM-CNN-based} \cite{xia2020lstm}. Similarly, for ECG-related datasets, comparisons are made with \textit{CNN-based models} \cite{kachuee2018ecg}.

Furthermore, we assess the hardware requirements for potential deployment on MCU environments. To facilitate fair comparisons, we re-implement these SOTA models, training them on the datasets used in this study. Additionally, we optimize these models using \textit{TensorFlow Lite} and profile their resource requirements using the \textit{MLTK} library to ensure suitability for deployment in resource-constraint environments.

Each dataset, along with the architectures produced by TinyTNAS and the re-implemented SOTA models, undergoes latency evaluation after optimization on two widely used low-cost, low-power IoT-enabled MCUs: \textit{ESP32} and \textit{Nano BLE Sense}. Here are their specifications:

\textbf{ESP32}: Dual-core Xtensa 32-bit LX6 microprocessors, 520 KB SRAM, 4 MB flash memory, priced at approximately \$5.

\textbf{Nano BLE Sense}: Nordic nRF52840 microcontroller, ARM Cortex-M4F processor, 256 KB RAM, 1 MB flash memory, priced at approximately \$30.

We measure the inference latency on each MCU, summarizing the results in Table \ref{ta1} for UCI HAR, Table \ref{ta2} for PAMAP2, Table \ref{ta3} for WISDM, Table \ref{ta4} for MIT-BIH, and Table \ref{ta5} for the PTB Diagnostics dataset. It is important to note that the inference time for \textit{Deep-Conv-LSTM-based} models \cite{ordonez2016deep} is not available due to high flash memory requirements, rendering the optimized models undeployable on the MCUs. These instances are marked as ND (Not Deployable) in the tables.

\begin{table}[!t]
\centering
\caption{Accuracy and Resource Requirements for Deploying Models Generated by TinyTNAS in TinyML Environments: RAM, FLASH, MAC,  Inference Time on Select MCUs, and Comparison with State-of-the-Art Methods for Dataset UCI-HAR}
\label{ta1}
\renewcommand{\arraystretch}{1.3} 
\resizebox{\textwidth}{!}{%
\begin{tabular}{c|c|c|c|c|c|c|c}
\hline
\multicolumn{1}{c|}{\textbf{Methods}} & \textbf{Structure} & \textbf{\begin{tabular}[c]{@{}c@{}}RAM\\ (kB)\end{tabular}} & \textbf{\begin{tabular}[c]{@{}c@{}}MAC\end{tabular}} & \textbf{\begin{tabular}[c]{@{}c@{}}FLASH\\(bytes)\end{tabular}} & \textbf{\begin{tabular}[c]{@{}c@{}}Accuracy\\ \%\end{tabular}} & \textbf{\begin{tabular}[c]{@{}c@{}}Latency (ms)\\ ESP32\end{tabular}} & \multicolumn{1}{c}{\textbf{\begin{tabular}[c]{@{}c@{}}Latency (ms)\\ Nano 33 BLE\end{tabular}}} \\ \hline
\cite{yang2015deep}
 & CNN (1d)& 53.6 &	1M&	623.9K&	92.93&	697	& 2525 \\
\cite{ordonez2016deep} & DeepConvLSTM (1d) & 124.4	&7.6M&	1.5M&	92.61&	ND&	ND \\
\cite{xia2020lstm} & LSTM-CNN (1d) & 38.9&	1.3M&	204.9K&	93.21&	1338&	4617 \\
\textbf{TinyTNAS} & Depthwise Seperable CNN (1d) & \textbf{10.8}&	\textbf{53.1K}&	\textbf{19.3K}&	\textbf{93.4}&	\textbf{13}&	\textbf{31}\\ \hline
\end{tabular}%
}
\end{table}

\begin{table}[!t]
\centering
\caption{Accuracy and Resource Requirements for Deploying Models Generated by TinyTNAS in TinyML Environments: RAM, FLASH, MAC,  Inference Time on Select MCUs, and Comparison with State-of-the-Art Methods for Dataset PAMAP2}
\label{ta2}
\renewcommand{\arraystretch}{1.3} 
\resizebox{\textwidth}{!}{%
\begin{tabular}{c|c|c|c|c|c|c|c}
\hline
\multicolumn{1}{c|}{\textbf{Methods}} & \textbf{Structure} & \textbf{\begin{tabular}[c]{@{}c@{}}RAM\\ (kB)\end{tabular}} & \textbf{\begin{tabular}[c]{@{}c@{}}MAC\end{tabular}} & \textbf{\begin{tabular}[c]{@{}c@{}}FLASH\\(bytes)\end{tabular}} & \textbf{\begin{tabular}[c]{@{}c@{}}Accuracy\\ \%\end{tabular}} & \textbf{\begin{tabular}[c]{@{}c@{}}Latency (ms)\\ ESP32\end{tabular}} & \multicolumn{1}{c}{\textbf{\begin{tabular}[c]{@{}c@{}}Latency (ms)\\ Nano 33 BLE\end{tabular}}} \\ \hline
\cite{yang2015deep}
 & CNN (1d) & 18.4 & 217.2K & 140.9K & 95.66 & 115 & 564 \\
\cite{ordonez2016deep} & DeepConvLSTM (1d)  & 34.3 & 1.8 M & 1.3 M & 96.44 & ND & ND \\
\cite{xia2020lstm} & LSTM-CNN (1d) & 16.4 & 357.1K & 203.3K & 95.97 & 337 & 1273 \\
\textbf{TinyTNAS} & Depthwise Seperable CNN (1d)  & \textbf{5.9} & \textbf{44.9K} & \textbf{15.8K} & \textbf{96.7} & \textbf{8} & \textbf{19} \\ \hline
\end{tabular}%
}
\end{table}

\begin{table}[!t]
\centering
\caption{Accuracy and Resource Requirements for Deploying Models Generated by TinyTNAS in TinyML Environments: RAM, FLASH, MAC,  Inference Time on Select MCUs, and Comparison with SOTA Methods for Dataset WISDM}
\label{ta3}
\renewcommand{\arraystretch}{1.3} 
\resizebox{\textwidth}{!}{%
\begin{tabular}{c|c|c|c|c|c|c|c}
\hline
\multicolumn{1}{c|}{\textbf{Algorithms}} & \textbf{Structure} & \textbf{\begin{tabular}[c]{@{}c@{}}RAM\\ (kB)\end{tabular}} & \textbf{\begin{tabular}[c]{@{}c@{}}MAC\end{tabular}} & \textbf{\begin{tabular}[c]{@{}c@{}}FLASH\\(bytes)\end{tabular}} & \textbf{\begin{tabular}[c]{@{}c@{}}Accuracy\\ \%\end{tabular}} & \textbf{\begin{tabular}[c]{@{}c@{}}Latency (ms)\\ ESP32\end{tabular}} & \multicolumn{1}{c}{\textbf{\begin{tabular}[c]{@{}c@{}}Latency (ms)\\ Nano 33 BLE\end{tabular}}} \\ \hline
\cite{yang2015deep}
 & CNN (1d) & 18.4 & 217.2K & 141.2K & 94.91 & 115 & 564 \\
\cite{ordonez2016deep} & DeepConvLSTM (1d) & 34.3 & 1.8 M & 1.3 M & 96.1 & ND & ND \\
\cite{xia2020lstm} & LSTM-CNN (1d) & 16.4 & 357.1K & 203.3K & 96.13 & 337 & 1273 \\
\textbf{TinyTNAS} &  Depthwise Seperable CNN (1d) & \textbf{5.9} & \textbf{44.9K} & \textbf{15.8K} & \textbf{96.5} & \textbf{8} & \textbf{19} \\ \hline
\end{tabular}%
}
\end{table}

\begin{table}[!t]
\centering
\caption{Accuracy and Resource Requirements for Deploying Models Generated by TinyTNAS in TinyML Environments: RAM, FLASH, MAC,  Inference Time on Select MCUs, and Comparison with State-of-the-Art Methods for Dataset MIT-BIH}
\label{ta4}
\renewcommand{\arraystretch}{1.3} 
\resizebox{\textwidth}{!}{%
\begin{tabular}{c|c|c|c|c|c|c|c}
\hline
\multicolumn{1}{c|}{\textbf{Algorithms}} & \textbf{Structure} & \textbf{\begin{tabular}[c]{@{}c@{}}RAM\\ (kB)\end{tabular}} & \textbf{\begin{tabular}[c]{@{}c@{}}MAC\end{tabular}} & \textbf{\begin{tabular}[c]{@{}c@{}}FLASH\\(bytes)\end{tabular}} & \textbf{\begin{tabular}[c]{@{}c@{}}Accuracy\\ \%\end{tabular}} & \textbf{\begin{tabular}[c]{@{}c@{}}Latency (ms)\\ ESP32\end{tabular}} & \multicolumn{1}{c}{\textbf{\begin{tabular}[c]{@{}c@{}}Latency (ms)\\ Nano 33 BLE\end{tabular}}} \\ \hline
\cite{kachuee2018ecg} & CNN (1d) & 80.7&	3.6M&	230.5K&	\textbf{98.36}&	2393&	9710\\
\textbf{TinyTNAS} & Depthwise Seperable CNN (1d) & \textbf{9}&	\textbf{56.5K}&	\textbf{19K}&	97.4&	\textbf{13}&	\textbf{33}\\ \hline
\end{tabular}%
}
\end{table}

\begin{table}[!t]
\centering
\caption{Accuracy and Resource Requirements for Deploying Models Generated by TinyTNAS in TinyML Environments: RAM, FLASH, MAC,  Inference Time on Select MCUs, and Comparison with State-of-the-Art Methods for Dataset PTB Diagnostic ECG Database }
\label{ta5}
\renewcommand{\arraystretch}{1.3} 
\resizebox{\textwidth}{!}{%
\begin{tabular}{c|c|c|c|c|c|c|c}
\hline
\multicolumn{1}{c|}{\textbf{Algorithms}} & \textbf{Structure} & \textbf{\begin{tabular}[c]{@{}c@{}}RAM\\ (kB)\end{tabular}} & \textbf{\begin{tabular}[c]{@{}c@{}}MAC\end{tabular}} & \textbf{\begin{tabular}[c]{@{}c@{}}FLASH\\(bytes)\end{tabular}} & \textbf{\begin{tabular}[c]{@{}c@{}}Accuracy\\ \%\end{tabular}} & \textbf{\begin{tabular}[c]{@{}c@{}}Latency (ms)\\ ESP32\end{tabular}} & \multicolumn{1}{c}{\textbf{\begin{tabular}[c]{@{}c@{}}Latency (ms)\\ Nano 33 BLE\end{tabular}}} \\ \hline
\cite{kachuee2018ecg} & CNN (1d) & 80.7&	3.6M&	230.5K&	\textbf{99.24}&	2393&	9710\\
\textbf{TinyTNAS} & Depthwise Seperable CNN (1d) & \textbf{9}&	\textbf{56.5K}&	\textbf{18.8K}&	95&	\textbf{13}&	\textbf{33}\\ \hline
\end{tabular}%
}
\end{table}

\subsection{Discussion}
Our evaluation highlights TinyTNAS's achievement of state-of-the-art (SOTA) accuracy while significantly reducing resource requirements—RAM, FLASH memory, latency and MAC operations—which are crucial for efficient time series classification tasks on TinyML domain. Given constraints of 20 KB RAM, 64 KB flash memory, 60K MAC operations, and a 10-minute search time on CPU, TinyTNAS sucessfuly generate architectures for given benchmark datasets.

Notably, TinyTNAS sets itself apart by employing a novel optimized grid search methodology instead of traditional reinforcement learning or evolutionary algorithms, optimizing architecture effectively on CPU-based but GPU-free systems within specified constraints. Detailed comparisons with alternative methods and comprehensive resource profiles are presented in Tables 
[\ref{ta1},\ref{ta2},\ref{ta3},\ref{ta4},\ref{ta5}].

In the case of the UCI HAR dataset, TinyTNAS-generated architectures meet the constraints and outperform other SOTA methods. It achieves a $12x$ reduction in RAM usage, $144x$ reduction in MAC operations, $78x$ reduction in flash memory, and a $149x$ reduction in latency. For the PAMAP2 and WISDM datasets, TinyTNAS achieves a $6x$ reduction in RAM usage, $40x$ reduction in MAC operations, $83x$ flash memory, and $67x$ reduction in latency, while maintaining superior accuracy compared to alternatives.

Moreover, for the PTB dataset, TinyTNAS achieves significant reductions: $9x$ less RAM usage, $64x$ fewer MAC operations, $13x$ lower flash memory requirement, and a $295x$ decrease in latency, with accuracy slightly below $5\%$. Similarly, for the MIT BIH dataset, TinyTNAS shows impressive reductions: $9x$ less RAM usage, $64x$ fewer MAC operations, $13x$ lower flash memory requirement, and a $295x$ decrease in latency, although its accuracy is slightly below $1\%$, indicating a minimal trade-off compared to other methods.

\section{Conclusion}
TinyTNAS represents a pioneering effort in bridging Neural Architecture Search (NAS) with TinyML, specifically targeting time series classification on resource-constrained devices. In this context, we introduce TinyTNAS, a novel hardware-aware multi-objective NAS tool designed specifically for TinyML time series classification. Unlike existing methods, TinyTNAS operates efficiently on CPUs without the need for GPUs, making it accessible and practical for a broader range of applications.

Users can define constraints on RAM, FLASH, and MAC operations to discover optimal neural network architectures that meet these parameters for a given dataset. Additionally, TinyTNAS allows for time-bound searches, ensuring the best possible model among the explored models is found within a user-specified duration. By leveraging an optimized grid search methodology that skips cells in a geometric progression in its explorable dimensions, TinyTNAS achieves state-of-the-art accuracy while drastically reducing RAM, FLASH memory, latency, and MAC operations, all within a very short search time on CPU.

Our comprehensive evaluations across benchmark datasets—UCI HAR, PAMAP2, WISDM, MIT-BIH, and PTB—demonstrate significant performance improvements compared to existing methods. These results underscore the effectiveness of integrating hardware-aware, multi-objective NAS approaches to efficiently meet stringent TinyML constraints. 

Importantly, to the best of our knowledge, this work represents the first comprehensive effort in developing a NAS tool specifically tailored for TinyML time series classification. It incorporates hardware-aware multi-objective optimization with constraints on RAM, MAC operations, FLASH memory, and time-bound searches, operating efficiently on CPUs without the need for GPUs. This sets a new standard in optimizing neural network architectures for AIoT and low-cost, low-power embedded AI applications.

\bibliographystyle{splncs04}
\bibliography{sample}

\end{document}